\UseRawInputEncoding
\documentclass [11pt] {article}
\usepackage{latexsym}
\usepackage{graphicx}

\RequirePackage{natbib}
\RequirePackage{url}
\usepackage{bm}

\newtheorem{theorem}{Theorem}

\title{Online Aggregation of Probability Forecasts with Confidence}
\author{{Vladimir V'yugin}
\\
{\small Institute for Information Transmission Problems}
\\
{\small (Moscow, Russia)}
\\
{\small e-mail vyugin@iitp.ru}
\\
{Vladimir Trunov}
\\
{\small Institute for Information Transmission Problems}
\\
{\small (Moscow, Russia)}
\\
{\small e-mail trunov@iitp.ru}}

\date{}

\newtheorem{lemma}{Lemma}

\def\argmin[{{\rm argmin}}

\def\y{{\bf y}}

\def\p{{\bf p}}
\def\q{{\bf q}}

\def\w{{\bf w}}

\def\0{{\bf 0}}
\def\c{{\bf c}}

\def\e{{\bf e}}

\def\Subst{{\rm Subst}}

\def\0{{\bf 0}}
\def\f{{\bf f}}

\def\CRPS{{\rm CRPS}}

\begin{document}

\maketitle





\begin{abstract}%
The paper presents numerical experiments and some
theoretical developments in prediction with expert advice (PEA). One experiment
deals with predicting electricity consumption depending on temperature and uses
real data. As the pattern of dependence can change with season and time of the day,
the domain naturally admits PEA formulation with experts having different
``areas of expertise''.
We consider the case where several competing methods produce online predictions
in the form of probability distribution functions.
The dissimilarity between a probability forecast and an outcome is measured by
a loss function (scoring rule). A popular example of scoring rule for continuous
outcomes is Continuous Ranked Probability Score ($\CRPS$).
In this paper the problem of combining probabilistic forecasts is considered in
the PEA framework. We show that $\CRPS$ is a mixable
loss function and then the time-independent upper bound for the regret of the
Vovk aggregating algorithm using $\CRPS$ as a loss function can be
obtained. Also, we incorporate a ``smooth'' version of the method of
specialized experts in this scheme which allows us to
combine the probabilistic predictions of the specialized experts with overlapping domains of
their competence.
\end{abstract}



\section{Introduction}

Probabilistic forecasts in the form of probability distributions over future events have
become popular in several fields, including meteorology, hydrology, economics,
demography.  
Probabilistic predictions are used in the theory of conformal predictions, where
a predictive distribution that is valid under a nonparametric assumption
can be assigned to any forecasting algorithm (see~\citet{VSMX2018}).

The dissimilarity between a probability forecast and an outcome is measured by
a loss function (scoring rule). A popular example of scoring rule for continuous
outcomes is Continuous Ranked Probability Score ($\CRPS$).
$$
\CRPS(F,y)=\int (F(u)-H(u-y))^2 du,
$$
where $F(u)$ is a probability distribution function, $y$ is an outcome -- a real
number, and $H(x)$ is the Heaviside function: $H(x)=0$ for $x<0$ and $H(x)=1$
for $x\ge 0$ (\citet{Eps69},~\citet{GnR2007}).

The paper presents theoretical developments in prediction with expert advice (PEA)
and some numerical experiments. One experiment
deals with predicting electricity consumption depending on temperature and uses
real data. As the pattern of dependence can change with season and time of the day,
the domain naturally admits PEA formulation with experts having a different
``areas of expertise''.

We consider the case where several competing methods produce online predictions
in the form of probability distribution functions. These predictions can lead to
large or small losses. Our task is to combine these forecasts into one optimal forecast,
which will lead to a relatively small possible loss in the framework of the
available past information.

We solve this problem in the PEA framework. We consider
the game-theoretic on-line learning model in which a learner (aggregating)
algorithm has to combine predictions from a set of $N$ experts
(see e.g.~\citet{LiW94},~\citet{FrS97},~\citet{Vov90},~\citet{KiW99},
~\citet{VoV98},~\citet{cesa-bianchi} among others).

In contrast to the standard PEA approach, we consider the case where each expert
presents probability distribution functions rather than a point prediction.
The learner presents his forecast also in the form of probability distribution function
computed using probabilistic predictions presented by the experts.

In online setting, at each time step $t$ each expert issues
a probability distribution as a forecast. The aggregating algorithm combines these
forecasts into one aggregated forecast, which is a probability distribution function.
The effectiveness of the aggregating algorithm
on any time interval $[1,T]$ is measured by the regret which is the difference between
the accumulated loss of the aggregating algorithm and the accumulated
loss of the best expert suffered on first $T$ steps.

There are many papers on probabilistic predictions and on $\CRPS$ scoring rule
(some of them are \citet{Bri1950}, \citet{Bro2007}, \citet{Bro2008}, \citet{Bro2012},
\citet{Eps69}, \citet{RGBP2005}).
In some cases, experts use for their predictions probability distributions functions
(data models) which are defined explicitly in an analytic form. In this paper, we propose
the rules for aggregation of such probability distributions functions.
We present the exact formulas for direct calculation of the aggregated probability
distribution function given probability distribution functions are presented by the experts.

We obtain a tight upper bound of the regret for a special
case when the outcomes and the probability distributions are supported on a finite
interval $[a,b]$ of real line.
In Section~\ref{main-1} we prove that the $\CRPS$ function is mixable and then
all machinery of the aggregating algorithm (AA) by~\citet{VoV98} and of the exponentially
weighted average forecaster (WA) (see~\citet{cesa-bianchi}) can be applied.
We present a method for computing online the aggregated probability distribution
function given the probability distribution functions are presented by the experts
and prove a time-independent bound for the regret of the proposed algorithm.

The application we will consider below in Section~\ref{exp-1}
(which is the sequential forecasting of
probability distribution function of electricity consumption) will take place in a variant
of the basic problem of prediction with expert advice called prediction with specialized
(or sleeping) experts. At each round, only some of the experts output a prediction while
the other ones are inactive. Each expert is expected to provide accurate forecasts
mostly under given external conditions that can be known beforehand. For instance,
in the case of the prediction of electricity consumption, experts can be specialized
to a season, temperature forecast, and time of the day.

Each expert is trained on its specific domain.
Moving from one domain to another, an expert which was tuned to the previous
domain gradually loses his predictive ability. To take this into account,
we define a smooth extension of the domain of any expert.
Thus, each expert competes with other experts working at overlapping intervals.
The second contribution of this paper is that we have incorporated a smooth
generalization of the method of specialized experts
(Sections~\ref{v-s-1} and~\ref{CRPS-3}) which allows us to combine the probabilistic predictions into
the aggregating algorithm (AA) of the specialized experts with overlapping domains of theirs competence.

We demonstrate the effectiveness of the proposed methods in Section~\ref{exp-1}, where
the results of numerical experiments with synthetic and real data are presented.

\section{Preliminaries}\label{app-1}

In this section we present the main definitions and the auxiliary results
of the theory of prediction with expert advice, namely, learning with mixable
loss functions.

\subsection{Online learning}\label{online-learning-1}

Let $\Omega$ be a set of outcomes and $\Gamma$ be a set of forecasts
(decision space).\footnote{In general, these sets can be of arbitrary nature.
We will specify them when necessary.}
We consider the learning with a loss function $\lambda(f,y)$, where
$f\in\Gamma$ and $y\in\Omega$.
Let also, a set $E$ of experts be given. For simplicity, we assume that $E=\{1,\dots ,N\}$.

In PEA approach the learning process is represented as a game. The experts and the learner
observe past real outcomes generated online by some adversarial mechanism (called nature)
and present their forecasts. After that, a current outcome is revealed by the nature.

In more detail, at any round $t=1,2,\dots$, each expert $i\in E$ presents a forecast $f_{i,t}\in\Gamma$,
then the learner presents its forecast $f_t\in\Gamma$, and after that, an outcome $y_t\in\Omega$
is revealed. Each expert $i$ suffers the loss $\lambda(f_{i,t},y_t)$, and the learner
suffers the loss $\lambda(f_t,y_t)$. The game of prediction with expert advice is presented by Protocol~1 below.

\smallskip

{\bf Protocol~1}

{\small
\medskip\hrule\hrule\medskip

\medskip
\noindent{\bf FOR} $t=1,\dots ,T$
\begin{enumerate}
\item Receive the experts' predictions $f_{i,t}$, where $1\le i\le N$.
\item Present the learner's forecast $f_t$.
\item Observe the true outcome $y_t$ and compute the losses
$\lambda(f_{i,t},y_t)$ of the experts and the loss $\lambda(f_t,y_t)$ of the learner.
\end{enumerate}

\noindent \hspace{2mm}{\bf ENDFOR}
\medskip\hrule\hrule\medskip
}
\smallskip

Let $H_T=\sum\limits_{t=1}^T \lambda(f_t,y_t)$ be the accumulated loss of the learner and
$L^i_T=\sum\limits_{t=1}^T \lambda(f_{i,t},y_t)$ be the accumulated loss of an expert $i$.
The difference $R^i_T=H_T-L^i_T$ is called
regret with respect to an expert $i$, and $R_T=H_T-\min_i L^i_T$ is the regret
with respect to the best expert. The goal of the learner is to minimize regret.

\subsection{Aggregating Algorithm (AA)}\label{AA-1}

The Vovk Aggregating algorithm (\citet{Vov90} and~\citet{VoV98}) is the base
algorithm for computing the learner predictions. This algorithm starting from
the initial weights $w_{i,1}$ (usually $w_{i,1}=\frac{1}{N}$ for all $i$)
assign weights $w_{i,t}$ for the experts $i\in E$ using the weights
update rule:
\begin{eqnarray}
w_{i,t+1}=w_{i,t}e^{-\eta\lambda(f_{i,t},y_t)}\mbox{ for } t=1,2,\dots,
\label{wei-up-1}
\end{eqnarray}
where $\eta>0$ is a learning rate. The normalized weights are defined
\begin{eqnarray}
w^*_{i,t}=\frac{w_{i,t}}{\sum\limits_{j=1}^N w_{j,t}}.
\label{weight-update-1}
\end{eqnarray}
The main tool of AA is a superprediction function
\begin{eqnarray}
g_t(y)=-\frac{1}{\eta}\ln\sum\limits_{i=1}^N e^{-\eta\lambda(f_{i,t},y)}w^*_{i,t}.
\label{superpred-1}
\end{eqnarray}
We consider probability distributions $\q=(q_1,\dots ,q_N)$ on
the set $E$ of the experts: $\sum\limits_{i=1}^N q_i=1$ and $q_i\ge 0$ for all $i$.
By~\citet{VoV98} a loss function is called $\eta$-mixable if for any probability
distribution $\q$ on the set $E$ of experts and for any predictions
$\f=(f_1,\dots ,f_N)$ of the experts there exists a forecast $f$ such that
\begin{eqnarray}
\lambda(f,y)\le g(y)\mbox{ for all } y,
\label{mix-1}
\end{eqnarray}
where
\begin{eqnarray}
g(y)=-\frac{1}{\eta}\ln\sum\limits_{i=1}^N e^{-\eta\lambda(f_i,y)}q_i.
\label{superpred-1a}
\end{eqnarray}
We fix some rule for calculating a forecast $f$ and write
\begin{eqnarray}
f=\Subst(\f,\q).
\label{AA_rule-1}
\end{eqnarray}
The function $\Subst$ is called the substitution function.

As follows from (\ref{mix-1}) and (\ref{superpred-1a}), if a loss function
$\lambda(f,y)$ is $\eta$-mixable, then the loss function
$c\lambda(f,y)$ is $\frac{\eta}{c}$-mixable for any $c>0$.

The upper bound
$H_T\le\sum\limits_{t=1}^T g_t(y_t)
\le L^i_T+\frac{\ln N}{\eta}$
for any expert $i$ is obtained in \ref{regret-ana-1}.
Therefore, there is a strategy for the learner that guarantees
the time-independent upper bound for the regret
$R_T\le\frac{\ln N}{\eta}$ for all $T$ regardless of which sequence
of outcomes is observed.

\subsection{Exponentially concave loss functions}\label{exp-conc-1}
Assume that all forecasts form a linear space. In this case,
the mixability is a generalization of the notion of exponentially concavity.
A loss function $\lambda(f,y)$ is called $\eta$-exponentially concave if
for each $y$ the function $\exp (-\eta\lambda(f,y))$ is concave in
$f$ (see~\citet{KiW99},~\citet{cesa-bianchi}). For exponentially concave loss function
the inequality (\ref{mix-1}) holds for all $y$ by definition if the
forecast of the learner is computed using the weighted average (WA)
of the experts predictions:
\begin{eqnarray}
f=\sum\limits_{i=1}^N q_i f_i,
\label{mix-forecast-2}
\end{eqnarray}
where $\q=(q_1,\dots ,q_N)$ is a probability distribution on the set of experts, and
$f_1,\dots ,f_N$ are theirs forecasts.

For exponentially concave loss function and the game defined by Protocol 1,
where the learner's forecast is computed by (\ref{mix-forecast-2}), we also have
the time-independent bound (\ref{prop-1}) for the regret.

\subsection{Square loss function} The important special case is $\Omega=\{0,1\}$
and $\Gamma=[0,1]$. The square loss function
$\lambda(\gamma,\omega)=(\gamma-\omega)^2$ is $\eta$-mixable loss function
for any $0<\eta\le 2$, where $\gamma\in [0,1]$ and $\omega\in\{0,1\}$.\footnote{
In what follows $\omega_t$ denotes a binary outcome.}
In this case, at any step $t$, the corresponding forecast $f_t$ (in Protocol 1)
can be defined as
\begin{eqnarray}
f_t=\Subst(\f_t,\w^*_t)=\frac{1}{2}-
\frac{1}{2\eta}\ln\frac{\sum\limits_{i=1}^N w^*_{i,t} e^{-\eta\lambda(f_{i,t},0)}}
{\sum\limits_{i=1}^N w^*_{i,t}e^{-\eta\lambda(f_{i,t},1)}},
\label{mix-forecast-1}
\end{eqnarray}
where $\f_t=(f_{1,t},\dots ,f_{N,t})$ is the vector of the experts' forecasts and
$\w^*_t=(w^*_{1,t},\dots ,w^*_{N,t})$ is the vector of theirs normalized weights
defined by (\ref{wei-up-1}) and~(\ref {weight-update-1}).
We refer the reader for details to \citet{Vov90},~\citet{VoV98}, and~\citet{Vov2001}.

The square loss function $\lambda(f,\omega)=(f-\omega)^2$ is $\eta$-exponential
concave for any $0<\eta\le\frac{1}{2}$ (see~\citet{cesa-bianchi}).

Note that the larger the learning rate, the faster the weights update rule
(\ref{wei-up-1}) adapts to the changing predictive abilities of the experts.

\section{AA for experts with confidence}\label{v-s-1}

In the experiments, which will be presented below in Section~\ref{exper-2},
the specialized experts will be used, where each expert is associated
with specific type of domain (time interval).

We define a smooth extension of the domain of any expert.
The scope of each expert will be determined by its confidence values.
Inside the area for which the expert was tuned, its confidence values are equal
to 1, and outside this area they decrease with time linearly from 1 to 0.

The method of specialized experts was first proposed by~\citet{Freu-2} and
further developed by~\citet{ChV2009},~\citet{electricity},~\citet{Gal2014},~\citet{KAS2015}.
With this approach, at each step $t$, a set of specialized experts
$E_t\subseteq\{1,\dots, N\}$ be given. A specialized expert $i$ issues its forecasts
not at all steps $t=1,2,\dots$, but only when $i\in E_t$. At any step, the aggregating
algorithm uses forecasts of only ``active (non-sleeping)'' experts.

We consider a more general case. At each time moment $t$, any expert's forecast
$f_{i,t}$ is supplied by a confidence level which is a real number $p_{i,t}\in [0,1]$.

In particular, $p_{i,t}=1$ means that the forecast of the expert $i$ is used in full,
whereas in the case of $p_{i,t}=0$ it is not taken into account at all (the expert sleeps).
In cases where $0<p_{i,t}<1$, the expert's forecast is partially taken into account.
For example, when moving from one season to another, an expert tuned to the previous
season gradually loses his predictive ability.
Confidence value can be set by the expert itself or by the learner.

The dependence of $p_{i,t} $ on values of exogenous parameters can be predetermined
by a specialist in the domain or can be constructed using regression analysis on
historical data.

The setting of prediction with experts that use confidence values as numbers
in the interval $[0,1]$ was studied (for Hedge algorithm)
by~\citet{BlM2007} and~\citet{Gal2014}. We modify this approach for AA algorithm.

Let $\lambda(f,y)$ be an $\eta$-mixable loss function.
At each time moment $t$ the forecasts $\f_t= (f_{1,t},\dots f_{N,t})$ of the experts
and confidence levels $\p_t=(p_{1,t},\dots ,p_{N,t})$ of these forecasts are
revealed.

In this section we modify AA for the experts with confidence.

To take into account confidence levels, we use the fixed point method by~\citet{ChV2009}.
We associate with any confidence level $p_{i,t}$ a probability distribution
$\p_{i,t}=(p_{i,t}, 1-p_{i,t})$ on a two element set.
Define the auxiliary probabilistic forecast:
\[
\tilde f_{i,t}=
\left\{
    \begin{array}{l}
      f_{i,t}\mbox{ with probability } p_{i,t},
    \\
      f_t\mbox{ with probability } 1-p_{i,t},
    \end{array}
  \right.
\]
where $f_t$ is a forecast of the learner.

First, we provide a justification of the algorithm presented below.
Our goal is to define the forecast $f_t$ such that
\begin{eqnarray}
e^{-\eta\lambda(f_t,y)}\ge\sum_{i=1}^N
E_{\p_{i,t}}[e^{-\eta\lambda(\tilde f_{i,t},y)}]w^*_{i,t}
\label{cond-1a}
\end{eqnarray}
for each $y$, where $\w^*_t=(w^*_{1,t},\dots ,w^*_{N,t})$ is the vector
of normalized weights defined by (\ref{wei-up-1}) and~(\ref {weight-update-1}).

Here $E_{\p_{i,t}}$ is the mathematical expectation
with respect to the probability distribution $\p_{i,t}$.
We rewrite inequality (\ref{cond-1a}) in a more detailed form:
\begin{eqnarray}
e^{-\eta\lambda(f_t,y)}\ge
\sum_{i=1}^N E_{\p_{i,t}}[e^{-\eta\lambda(\tilde f_{i,t},y)}]w^*_{i,t}=
\label{cond-1}
\\
\sum_{i=1}^N p_{i,t}w^*_{i,t}e^{-\eta\lambda(f_{i,t},y)}+
e^{-\eta\lambda(f_t,y)}
\left(1-\sum_{i=1}^N p_{i,t}w^*_{i,t}\right)
\label{cond-2}
\end{eqnarray}
for all $\omega$.
Therefore, the inequality (\ref{cond-1a}) is equivalent to the inequality
\begin{eqnarray}
e^{-\eta\lambda(f_t,y)}\ge\sum_{i=1}^N
w^p_{i,t}e^{-\eta\lambda(f_{i,t},y)},
\label{for-1ba}
\end{eqnarray}
where
\begin{eqnarray}\label{for-1bb}
w^p_{i,t}=\frac{p_{i,t} w^*_{i,t}}{\sum_{j=1}^N p_{j,t} w^*_{j,t}}=\frac{p_{i,t} w_{i,t}}{\sum_{j=1}^N p_{j,t} w_{j,t}}.
\end{eqnarray}
According to the rule (\ref{AA_rule-1}) for computing the forecast of AA,
define $f_t=\Subst(\f_t,\w^p_t)$. Then
(\ref{for-1ba}) and its equivalent (\ref{cond-1}) are valid.
Here $\Subst$ is the substitution function, $\w^p_t=(w^p_{i,1},\dots , w^p_{i,N})$ and
$\f_t=(f_{1,t},\dots f_{i,N})$.
.
Let us refine Protocol 1 in the form of Algorithm 1a which is
the algorithm AA with confidence. This algorithm presents a strategy for the learner in Protocols 1.

\smallskip

{\bf Algorithm 1a}

{\small
\medskip\hrule\hrule\medskip

\medskip
\noindent{\bf FOR} $t=1,\dots ,T$
\begin{enumerate}
\item Receive the experts' predictions $f_{i,t}$ and confidence levels $p_{i,t}$,
where $1\le i\le N$.
\item Present the learner's forecast $f_t=\Subst(\f_t,\w^p_t)$, where
normalized weights $\w^p_t=(w^p_{1,t},\dots ,w^p_{N,t})$ are defined by (\ref{for-1bb}).
\item Observe the true outcome $y_t$ and compute the losses
$l_{i,t}=\lambda(f_{i,t},y_t)$ of the experts and the loss $\lambda(f_t,y_t)$ of the learner.
\item Update the weights (of the virtual experts) by the rule 
\begin{eqnarray}
w_{i,t+1}=w_{i,t}e^{-\eta (p_{i,t}\lambda(f_{i,t},y_t)+(1-p_{i,t})\lambda(f_t,y_t))}
\label{for-1b-2}
\end{eqnarray}
\end{enumerate}

\noindent \hspace{2mm}{\bf ENDFOR}
\medskip\hrule\hrule\medskip
}
\smallskip

Let $l_{i,t}=\lambda(f_{i,t},y_t)$ be the loss of an expert $i$ and
$h_t=\lambda(f_t,y_t)$ be the loss of the learner at step $t$.
Define the estimated loss of an expert $i$ as $\tilde l_{i,t}=\lambda(\tilde f_{i,t},y_t)$
and $\hat l_{i,t}=E_{\p_{i,t}}[\tilde l_{i,t}]$ be its expectation.
By the virtual expert $i$ we mean the expert which suffers the loss $\hat l_{i,t}$.

Since by definition $\hat l_{i,t}=p_{i,t}l_{i,t}+(1-p_{i,t})h_t$,
we have $h_t-\hat l_{i,t}=p_{i,t}(h_t-l_{i,t})$. We call the last quantity
discounted excess loss with respect to an expert $i$ at a time moment $t$ and
we will measure the performance of our algorithm by the cumulative discounted excess
loss with respect to any expert $i$.
\begin{theorem}\label{main-3fa}
For any ${1\le i\le N}$, the following upper bound
for the cumulative excess loss (discounted regret) holds true:
\begin{eqnarray}
\sum\limits_{t=1}^T p_{i,t}(h_t-l_{i,t})\le\frac{\ln N}{\eta}
\label{mTT-1afaa}
\end{eqnarray}
for all $T$.
\end{theorem}
{\it Proof}.
By convexity of the exponent the inequality (\ref{cond-1a}) implies
\begin{eqnarray}
e^{-\eta\lambda(f_t,y)}\ge\sum_{i=1}^N e^{-\eta E_{\p_{i,t}}
[\lambda(\tilde f_{i,t},y)]}w^*_{i,t}=
\sum_{i=1}^N e^{-\eta\hat l_{i,t}}w^*_{i,t}.
\label{for-1b-2s}
\end{eqnarray}
Rewrite the update rule (\ref{for-1b-2}) as
$w_{i,t+1}=w_{i,t}e^{-\eta\hat l_{i,t}}$.
Using the regret analysis for AA in~\ref{regret-ana-1}, we obtain
\begin{eqnarray*}
\sum_{t=1}^T h_t\le
\sum_{t=1}^T \hat l_{i,t}+\frac{\ln N}{\eta}
\end{eqnarray*}
for any $i$.
Since $h_t-\hat l_{i,t}=p_{i,t}(h_t-l_{i,t})$, the inequality (\ref{mTT-1afaa})
follows.
$\Box$

\section{Aggregation of probability forecasts}\label{main-1}


Let the set of outcomes in Protocol 1 be an interval $\Omega=[a,b]$ of the real line
for some $a<b$ and the set of forecasts $\Gamma$ be a set of all
probability distribution functions $F:[a,b]\to [0,1]$.\footnote{
A probability distribution function is a non-decreasing function
$F(y)$ defined on this interval such that $F(a)=0$ and $F(b)=1$.
Also, it is right-continuous and has the left limit at each point.}

The quality of the prediction $F$ in view of the actual outcome $y$
is often measured by the continuous ranked probability score (loss function)
\begin{equation}
\CRPS(F,y)=\int_a^b (F(u)-H(u-y))^2 du,
\label{crps-1}
\end{equation}
where $H(x)$ is the Heaviside function: $H(x)=0$ for $x<0$ and $H(x)=1$ for $x\ge 0$
(\citet{Eps69},~\citet{MaW76}, etc).

For simplicity, we consider in this definition integration over a finite
interval.
Such definition is closer to practical applications and
allows a more elementary theoretical analysis.
More general definition includes a density $\mu(u)$ and integration over the real line:
\begin{equation}
\CRPS(F,y)=\int_{-\infty}^{+\infty} (F(u)-H(u-y))^2 \mu(u)du.
\label{crps-1a}
\end{equation}
The definition (\ref{crps-1}) is a special case of this definition (up to a factor),
where $\mu(u)=\frac{1}{b-a}$ for $u\in [a,b]$ and $\mu(u)=0$ otherwise.
It can be proved that the function (\ref{crps-1a}) is $\eta$-mixable for $0<\eta\le 2$
and $\eta$-exponentially concave for $0<\eta\le\frac{1}{2}$ (see~\citet{KVG2019}).

The $\CRPS$ score measures the difference between the forecast $F$ and a perfect forecast
$H(u-y)$ which puts all mass on the verification $y$.
The lowest possible value $0$ is attained
when $F$ is concentrated at $y$, and in all other cases $\CRPS(F,y)$ will be positive.

We consider a game of prediction with expert advice, where the forecasts of the experts
and of the learner are (cumulative) probability distribution functions.
At any step $t$ of the game
each expert $i\in \{1,\dots ,N\}$ presents its forecast -- a probability distribution
function $F_{i,t}(u)$ and the learner presents its forecast $F_t(u)$.\footnote{
For simplicity of presentation, we consider the case where the set of the experts is finite.
In case of infinite $E$, the sums by $i$ should be replaced by integrals with respect
to the corresponding probability distributions on the set of experts. In this case
the choice of initial weights on the set of the experts is a non-trivial problem.}
After an outcome $y_t\in [a,b]$ have been revealed and the experts and the learner
suffer losses $\CRPS(F_{i,t},y_t)$ and $\CRPS(F_t,y_t)$.

The corresponding game of probabilistic prediction is defined by the following protocol.

\smallskip

{\bf Protocol~2}

{\small
\medskip\hrule\hrule\medskip

\medskip
\noindent{\bf FOR} $t=1,\dots ,T$
\begin{enumerate}
\item Receive the experts' predictions -- the probability distribution
functions $F_{i,t}(u)$ for $1\le i\le N$.
\item Present the learner's forecast -- the probability distribution
function $F_t(u)$.
\item Observe the true outcome $y_t$ and compute the scores

$\CRPS(F_{i,t},y_t)=\int_a^b (F_{i,t}(u)-H(u-y_t))^2 du$
of the experts $1\le i\le N$

and the score

$\CRPS(F_t,y_t)=\int_a^b (F_t(u)-H(u-y_t))^2 du$
of the learner.
\end{enumerate}

\noindent \hspace{2mm}{\bf ENDFOR}
\medskip\hrule\hrule\medskip
}
\smallskip

The goal of the learner is to predict in such a way that independently
of which outcomes
are revealed and the experts' predictions are presented, its accumulated loss
$H_T=\sum\limits_{t=1}^T\CRPS(F_t,y_t)$ is asymptotically less
than the loss $L^i_T=\sum\limits_{t=1}^T\CRPS(F_{i,t},y_t)$
of the best expert $i$ up to some regret and $H_T-\min_i L^i_T=o(T)$ as $T\to\infty$.

First, we show that $\CRPS$ loss function (and the corresponding game) is mixable.
\begin{theorem}\label{theorem-1}
The continuous ranked probability score $\CRPS(F,y)$ is $\frac{2}{b-a}$-mixable
loss function.
The corresponding learner's forecast $F(u)$ given the forecasts $F_i(u)$ of
the experts $1\le i\le N$ and a probability distribution $\q=(q_1,\dots ,q_N)$ on
the set of all experts can be computed by the rule \footnote{
Ii is easy to verify that $F(u)$ is a probability distribution function.}
\begin{eqnarray}
F(u)=\frac{1}{2}-\frac{1}{4}\ln\frac{\sum_{i=1}^N q_i e^{-2(F_i(u))^2}}
{\sum_{i=1}^N q_i e^{-2(1-F_i(u))^2}},
\label{forecast-2}
\end{eqnarray}
\end{theorem}
{\it Proof}.
We approximate any probability distribution function $F(u)$ by
a piecewise-constant function that takes a finite number
of values on a uniform grid of arguments. Accordingly, the forecasts of the experts and
of the learner will take the form of $d$-dimensional vectors, where $d$ is a positive
integer number. We apply AA to the $d$-dimensional forecasts, then we consider
the limit $d\to\infty$.

~\citet{Kaln2017} generalize the AA for the case of $d$-dimensional forecasts, where
$d$ is a positive integer number.
Let an $\eta$-mixable loss function $\lambda(f,y)$ be given, where $\eta>0$, $f\in\Gamma$
and $y\in\Omega$. Let $\f=(f^1,\dots ,f^d)\in\Gamma^d$ be a $d$-dimensional forecast and
$\y=(y^1,\dots ,y^d)\in\Omega^d$ be a $d$-dimensional outcome. The generalized loss
function is defined $\lambda(\f,\y)=\sum\limits_{s=1}^d\lambda(f^s,y^s)$; we call
$\lambda(f,y)$ its source function.

The corresponding (generalized) game can be presented by Protocol 1
where at each step $t$ the experts
and the learner present $d$-dimensional forecasts: at any round $t=1,2,\dots$ each
expert $i\in\{1,\dots ,N\}$ presents a vector of forecasts
$\f_{i,t}=(f^1_{i,t},\dots, f^d_{i,t})$ and the learner
presents a vector of forecasts $\f_t=(f^1_t,\dots ,f^d_t)$.
After that, a vector $\y_t=(y^1_t,\dots ,y^d_t)$ of outcomes
will be revealed and the experts and the learner suffer losses
$\lambda(\f_{i,t},\y_t)=\sum\limits_{s=1}^d\lambda(f^s_{i,t},\y^s_t)$
and
$\lambda(\f_t,\y_t)=\sum\limits_{s=1}^d\lambda(f^s_t,y^s_t)$.

~\citet{Kaln2017} proved that the generalized loss function (game) is mixable.
\begin{lemma}\label{Adam-1}
The generalized loss function
$\lambda(\f,\y)$ is $\frac{\eta}{d}$-mixable if the source loss function
$\lambda(f,y)$ is $\eta$-mixable.
\end{lemma}
We reproduce the proof in~\ref{adam-2} for completeness of presentation.

We now turn to the proof of Theorem~\ref{theorem-1}.
We approximate any probability distribution
function $F(y)$ by piecewise-constant functions $F_d(y)$, where $d=1,2,\dots$.
Any such function $F_d$ is defined by the points $z_0,z_1,z_2,\dots ,z_d$ and
the values $f_0=F(z_0)$, $f_1=F(z_1)$, $f_2=F(z_2)$, $\dots$, $f_d=F(z_d)$,
where $a=z_0<z_1<z_2<\dots <z_d=b$ and
$0=f_0\le f_1\le f_2\le \dots \le f_d=1$. By definition
$F_d(y)=f_i$ for $z_{i-1}<y\le z_i$, where $1\le i\le d$.
Also, assume that $z_{i+1}-z_i=\Delta$ for all $0\le i<d$.
By definition $\Delta=\frac{b-a}{d}$.
Since $F(u)\le F_d(u)$ for all $u$,
\begin{eqnarray}
\left|\CRPS(F,y)-\CRPS(F_d,y)\right|\le
\nonumber
\\
\int_a^y (F_d^2(u)-F^2(u))du+
\int_y^b ((1-F(u))^2-(1-F_d(u))^2)du
\label{appr-2s}
\end{eqnarray}
for any $y\in [a,b]$. 
Let $z_{k-1}<y\le z_k$, where $1\le k\le d$. Then
\begin{eqnarray*}
\int_a^y (F_d^2(u)-F^2(u))du\le\sum_{i=0}^{k-1}\int_{z_i}^{z_{i+1}}(F_d^2(u)-F^2(u))du\le
\nonumber
\\
\Delta\sum_{i=0}^{k-1} F_d^2(z_{i+1})-F_d^2(z_i)=\Delta(F_d^2(z_k)-F_d^2(a))\le\Delta.
\end{eqnarray*}
The second integral in (\ref{appr-2s}) is also bounded by $\Delta$.
Hence,
\begin{equation}\label{appr-2}
\left|\CRPS(F,y)-\CRPS(F_d,y)\right|\le 2\Delta.
\end{equation}

Define an auxiliary representation of $y$, which is a binary variable
$\omega_{y,s}=1_{z_s\ge y}\in\{0,1\}$ for $1\le s\le d$ and
${\bm\omega}_y=(\omega_{y,1},\dots,\omega_{y,d})$, where $1_{z_s\ge y}=H(z_s-y)$.

Consider any $y\in [a,b]$. It is easy to see that for each $1\le s\le d$
the uniform measure of all $u\in [z_{s-1},z_s]$
such that $1_{z_s\ge y}\not =1_{u\ge y}$ is less or equal to $\Delta$ if
$y\in [z_{s-1},z_s]$ and $1_{z_s\ge y}=1_{u\ge y}$ for all $u\in [z_{s-1},z_s]$
otherwise. Since $0\le f_s\le 1$ for all $s$, this implies that
\begin{eqnarray}
\left|\CRPS(F_d,y)-\Delta\sum_{s=1}^d (f_s-\omega_{y,s})^2\right|= 
\nonumber
\\
\left|\int_{z_{k-1}}^{z_k}(f_k-1_{u\ge y})^2 du-\Delta (f_k-\omega_{y,k})^2\right|\le
\nonumber
\\
\Delta \left| f_k^2-(f_k-1)^2\right|=\Delta\left|2f_k-1\right|\le\Delta,
\label{appr-1}
\end{eqnarray}
where $y\in (z_{k-1},z_k]$. Let us study the generalized loss function
\begin{eqnarray}
\lambda(\f,{\bm\omega})=\Delta\sum_{s=1}^d (f_s-\omega_s)^2,
\label{vf-1}
\end{eqnarray}
where $\f=(f_1,\dots ,f_d)$, ${\bm\omega}=(\omega_1,\dots,\omega_d)$ and
$\omega_s\in\{0,1\}$ for $1\le s\le d$.

The key observation is that the deterioration of the learning rate for the
generalized loss function (it gets divided by the dimension $d$ of vector-valued forecasts)
is exactly offset by the decrease in the weight of each component of the vector-valued
prediction as the grid-size decreases.

Since the square loss function $\lambda(f,\omega)=(\gamma-\omega)^2$ is
$2$-mixable, where $f\in [0,1]$ and $\omega\in\{0,1\}$, by results of
Section~\ref{app-1} the corresponding generalized loss function
$\sum_{s=1}^d (f_s-\omega^s)^2$ is $\frac{2}{d}$-mixable and then the loss function
(\ref{vf-1}) is $\frac{2}{d\Delta}=\frac{2}{b-a}$-mixable independently of what
grid-size is used.\footnote{This also means that in numerical experiments,
when calculating forecasts of the learner, we can use the same learning rate,
regardless of the accuracy of the presentation of expert forecasts.}

Let $F_i(u)$ be the probability distribution functions presented by the experts
$1\le i\le N$ and $\f_i=(f_{i,1},\dots ,f_{i,d})$, where $f_{i,s}=F_i(z_s)$ for
$1\le s\le d$.
By (\ref{ii-1})
\begin{eqnarray}
e^{-\frac{2}{(b-a)}\lambda(\f,{\bm\omega})}\ge\sum_{i=1}^N
e^{-\frac{2}{b-a}\lambda(\f_i,{\bm\omega})}q_i
\label{ii-1k}
\end{eqnarray}
for each ${\bm\omega}\in\{0,1\}^d$ (including ${\bm\omega}={\bm\omega}_y$
for any $y\in [a,b]$), where the forecast $\f=(f_1,\dots ,f_d)$ can be defined as
\begin{eqnarray}
f_s=\frac{1}{2}-\frac{1}{4}\ln\frac{\sum_{i=1}^N q_i e^{-2(f_{i,s})^2}}
{\sum_{i=1}^N q_i e^{-2(1-f_{i,s})^2}}
\label{forecast-1f}
\end{eqnarray}
for each $1\le s\le d$.

By letting the grid-size $\Delta\to 0$ (or, equivalently, $d\to\infty$) in (\ref{appr-1}),
(\ref{ii-1k}), where ${\bm\omega}={\bm\omega}_y$, and in (\ref{appr-2}), we obtain
for any $y\in [a,b]$,
\begin{eqnarray}
e^{-\frac{2}{(b-a)}\CRPS(F,y)}\ge\sum_{i=1}^N e^{-\frac{2}{b-a}\CRPS(F_i,y)}q_i,
\label{ii-1kk}
\end{eqnarray}
where $F(u)$ is the limit form of (\ref{forecast-1f}) defined by
\begin{eqnarray*}
F(u)=\frac{1}{2}-\frac{1}{4}\ln\frac{\sum_{i=1}^N q_i e^{-2(F_i(u))^2}}
{\sum_{i=1}^N q_i e^{-2(1-F_i(u))^2}}
\end{eqnarray*}
for each $u\in [a,b]$.

The inequality (\ref{ii-1kk}) means that the loss function
$\CRPS(F,y)$ is $\frac{2}{b-a}$-mixable.
$\Box$

Let us refine the protocol 2 of the game with probabilistic predictions
for the case when the rule (\ref{forecast-2}) for AA is used.
This algorithm presents a strategy for the learner in Protocol 2.

\smallskip

{\bf Algorithm~3}

{\small
\medskip\hrule\hrule\medskip

Define $w_{i,1}=\frac{1}{N}$ for $1\le i\le N$.

\medskip
\noindent \hspace{2mm}{\bf FOR} $t=1,\dots ,T$
\begin{enumerate}
\item Receive the expert predictions -- the probability distribution
functions $F_{i,t}(u)$, where $1\le i\le N$.
\item Present the learner forecast -- the probability distribution
function $F_t(u)$:
\begin{eqnarray}
F_t(u)=\frac{1}{2}-\frac{1}{4}\ln\frac{\sum_{i=1}^N w^*_{i,t}e^{-2(F_{i,t}(u))^2}}
{\sum_{i=1}^N w^*_{i,t}e^{-2(1-F_{i,t}(u))^2}},
\label{forecast-2b}
\end{eqnarray}
where $w^*_{i,t}=\frac{w_{i,t}}{\sum_{j=1}^N w_{j,t}}$.
\item
Observe the true outcome $y_t$ and compute the score
$\CRPS(F_{i,t},y_t)$ for the experts $1\le i\le N$
and the score $\CRPS(F_t,y_t)$ for the learner.
\item
Update the weights of the experts $1\le i\le N$
\begin{eqnarray}
w_{i,t+1}=w_{i,t}e^{-\frac{2}{b-a}\CRPS(F_{i,t},y_t)}
\label{weight-up-3}
\end{eqnarray}
\end{enumerate}

\noindent \hspace{2mm}{\bf ENDFOR}
\medskip\hrule\hrule\medskip
}
\smallskip

The performance bound of Algorithm 3 is presented in the following theorem.

\begin{theorem}\label{theorem-2}
For each $T$,
\begin{eqnarray}
\sum\limits_{t=1}^T\CRPS(F_t,y_t)\le\min_{1\le i\le N}\sum\limits_{t=1}^T \CRPS(F_{i,t},y_t) +
\frac{b-a}{2}\ln N.
\label{h-t-2a}
\end{eqnarray}
\end{theorem}
{\it Proof.} The bound (\ref{h-t-2a}) is a direct corollary of the regret analysis
of
~\ref{regret-ana-1} and the bound (\ref{prop-1}).
$\Box$

The square loss function is also $\eta$-exponentially concave for $0<\eta\le\frac{1}{2}$
(see~\citet{cesa-bianchi}). In this case (\ref{forecast-2b}) can be replaced with
the forecast WA
\begin{eqnarray}
F_t(u)=\sum\limits_{i=1}^N w^*_{i,t} F_{i,t}(u),
\label{forecast-2a}
\end{eqnarray}
where $w^*_{i,t}=\frac{w_{i,t}}{\sum\limits_{j=1}^N w_{j,t}}$ are normalized weights.
The corresponding weights are computed recursively
\begin{eqnarray}
w_{i,t+1}=w_{i,t}e^{-\frac{1}{2(b-a)}\CRPS(F_{i,t},y_t)}.
\label{exp-concave-1}
\end{eqnarray}
Using Lemma~\ref{Adam-1} and Theorem~\ref{theorem-2}, we conclude that
in this case the bound (\ref{h-t-2a}) can be replaced with
\begin{eqnarray*}
\sum\limits_{t=1}^T {\CRPS}(F_t,y_t)\le\min_{1\le i\le N}
\sum\limits_{t=1}^T {\CRPS}(F_{i,t},y_t)+2(b-a)\ln N.
\end{eqnarray*}
The proof is similar to the proof of Theorem~\ref{theorem-2}.

\subsection{Aggregation of probabilistic predictions with confidence}\label{CRPS-3}

In Section~\ref{exper-2} (below), we present results of numerical experiments
with the real data and when probabilistic predictions of the experts are supplied with the
levels of confidence. In this case we use Algorithm 3a as a strategy of the learner,
that is a modification of Algorithm 3.

At each round, only some of the experts output a prediction while
the other ones are inactive. Each expert is expected to provide accurate forecasts
mostly under given external conditions that can be known beforehand, namely,
the experts are specialized to a season, temperature forecast, and time of the day.

We define a smooth extension of the domain of any expert.
Thus, each expert competes with other experts working at overlapping intervals.

The aggregating algorithms AA and WA allow us to combine the probabilistic predictions
of the specialized experts with overlapping domains of theirs competence.

\smallskip

{\bf Algorithm~3a (Strategy for the learner)}

{\small
\medskip\hrule\hrule\medskip

Define $w_{i,1}=\frac{1}{N}$ for $1\le i\le N$.

\medskip
\noindent \hspace{2mm}{\bf FOR} $t=1,\dots ,T$
\begin{enumerate}
\item Receive the expert predictions -- the probability distribution
functions $F_{i,t}(u)$ and confidence levels $p_{i,t}$, where $1\le i\le N$.
\item Present the learner forecast -- the probability distribution
function $F_t(u)$ which is defined by the rule
\begin{eqnarray}
F_t(u)=\frac{1}{2}-\frac{1}{4}\ln\frac{\sum_{i=1}^N w^p_{i,t}e^{-2(F_{i,t}(u))^2}}
{\sum_{i=1}^N w^p_{i,t}e^{-2(1-F_{i,t}(u))^2}}
\label{forecast-2bc}
\end{eqnarray}
for AA or by the rule
\begin{eqnarray}
F_t(u)=\sum\limits_{i=1}^N w^p_{i,t} F_{i,t}(u)
\label{forecast-2ac}
\end{eqnarray}
for WA, where
$w^p_{i,t}=\frac{p_{i,t} w_{i,t}}{\sum_{j=1}^N p_{j,t} w_{j,t}}$.
\item
Observe the true outcome $y_t$ and compute the score
$\CRPS(F_{i,t},y_t)$ 
for the experts $1\le i\le N$
and the score
$\CRPS(F_t,y_t)$ 
for the learner.
\item
Update the weights of the (virtual) experts $1\le i\le N$
\begin{eqnarray}
w_{i,t+1}=w_{i,t}e^{-\eta (p_{i,t}\CRPS(F_{i,t},y_t)+(1-p_{i,t})\CRPS(F_t,y_t))},
\label{for-1b-2c}
\end{eqnarray}
where $\eta=\frac{2}{b-a}$ for AA and 
$\eta=\frac{1}{2(b-a)}$ for WA. 
\end{enumerate}
\noindent \hspace{2mm}{\bf ENDFOR}
\medskip\hrule\hrule\medskip
}
\smallskip

The performance of the algorithm
is presented by the inequality (\ref{mTT-1afaa})
of Theorem~\ref{main-3fa}, where $h_t=\CRPS(F_t,y_t)$,
$l_{i,t}=\CRPS(F_{i,t},y_t)$ and $\eta=\frac{2}{b-a}$ if the rule (\ref{forecast-2bc})
for computing the learner's forecast was used and $\eta=\frac{1}{2(b-a)}$ if the rule
(\ref{forecast-2ac}) was used.

The proposed rules (\ref{forecast-2bc}) for AA and (\ref{forecast-2ac}) for WA can be used
when the probability distributions presented by the experts are
given in the closed form (i.e., distributions given by analytical formulas).
For this case, numerical methods can be used to calculate the integrals ($\CRPS$) with
any degree of accuracy given in advance (see also Footnote 6).

\section{Experiments}\label{exp-1}

In this section we apply our proposed algorithm on synthetic data and on electricity
consumption data, and compare its performance for several predictive models.
We use Algorithm 3 in the experiments with synthetic data and Algorithm 3a for
the electricity consumption data.

To optimize the losses in our mixing schemes, we
used the mixing past posteriors modification of Algorithms 3 and 3a,
see~\ref{mix-past-1}.

The algorithms and the data are presented at GitHub:
\url{https://github.com/VladimirVyugin},
Project ``Online-Aggregation-of-Probability-Forecasts -With-Confidence''

\subsection{Synthetic data}\label{exper-1}

\begin{figure}[!htb]
\includegraphics[scale=0.30]{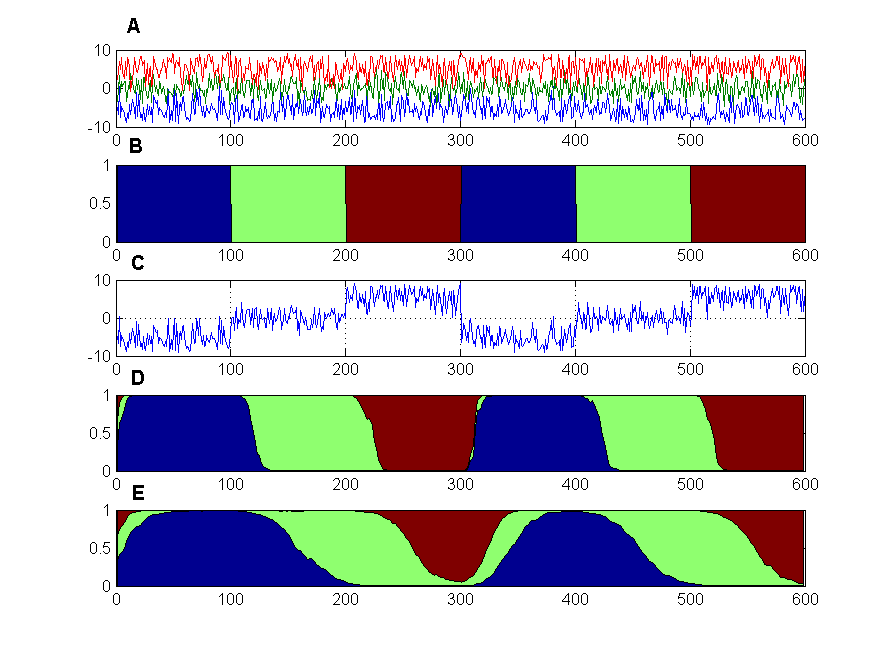}
\includegraphics[scale=0.30]{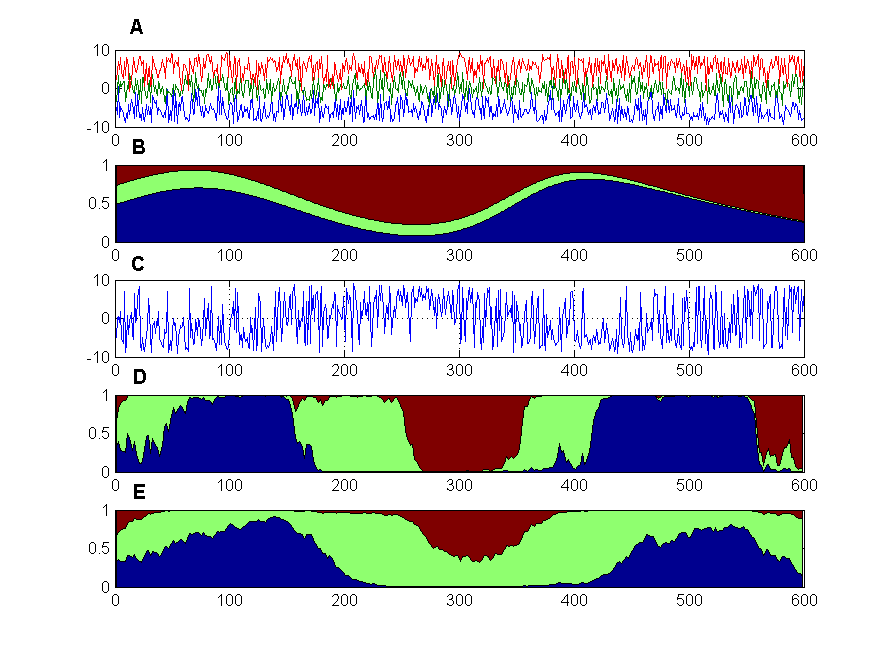}
        \caption{{\small The stages of numerical experiments and the results of
experts' aggregation for two initial synthetic data mixing methods (Method 1 -- left, Method 2 - right).
(A) Realizations of the trajectories for the three initial data generating distributions;
(B) weights of the distributions assigned by the data mixing method;
(C) sequence sampled from the distributions defined by Method 1 and Method 2;
(D) weights of the experts assigned online by AA using the rule
(\ref{weight-up-3}); (E) weights of the experts assigned online by WA using
the rule (\ref{exp-concave-1}).
}}\label{fig-2}
          \end{figure}

\begin{figure}[!htb]
\includegraphics[scale=0.30]{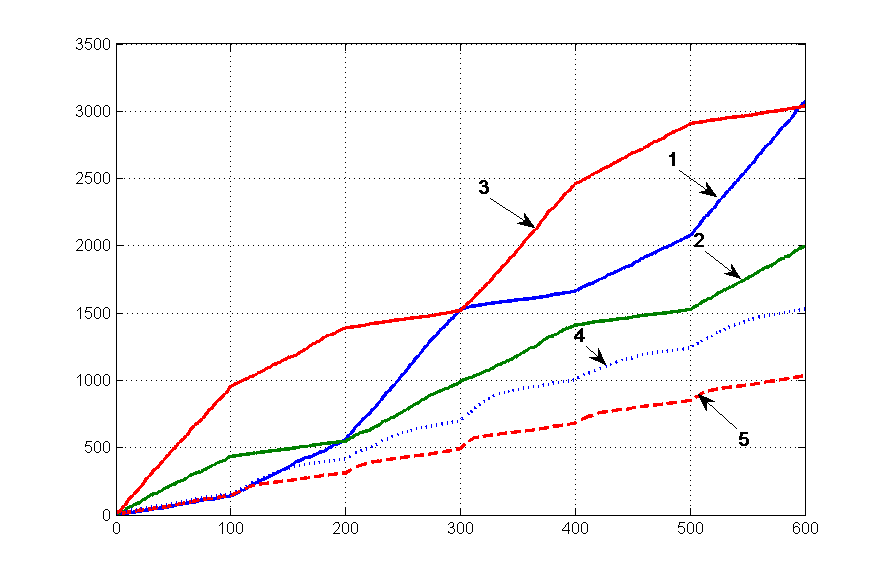}
\includegraphics[scale=0.30]{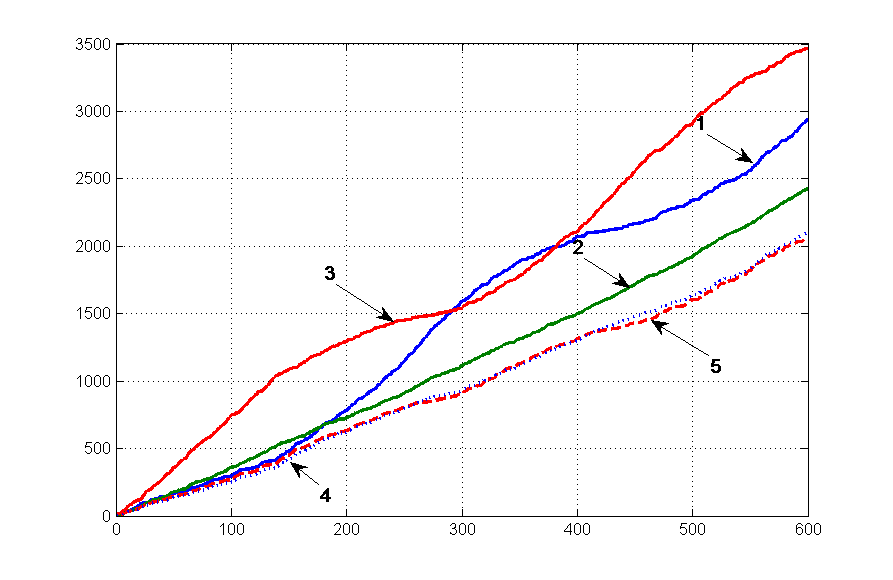}
        \caption{{\small
The accumulated losses of the experts (lines 1-3) and
of the aggregating algorithm for both initial data mixing methods
(Method 1 -- left, Method 2 - right) and for both methods of computing
aggregated forecasts:
line 4 -- for WA (the rule (\ref{forecast-2a})) and line 5 -- for AA (the rule
(\ref{forecast-2b})). We note an advantage of AA 
over WA
in the case of data generating Method 1,
in which there is a rapid change in leadership of the data generating distributions.
}}\label{fig-3}
                  \end{figure}

\begin{figure}[!htb]
\includegraphics[scale=0.30]{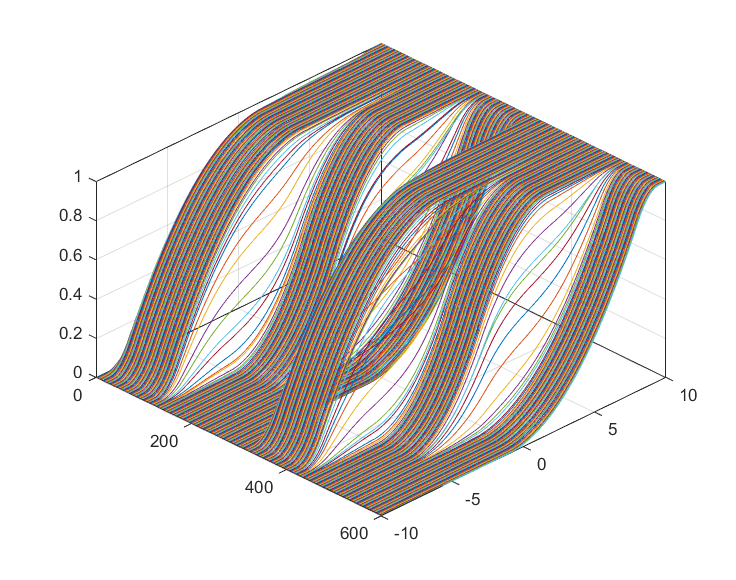}
\includegraphics[scale=0.30]{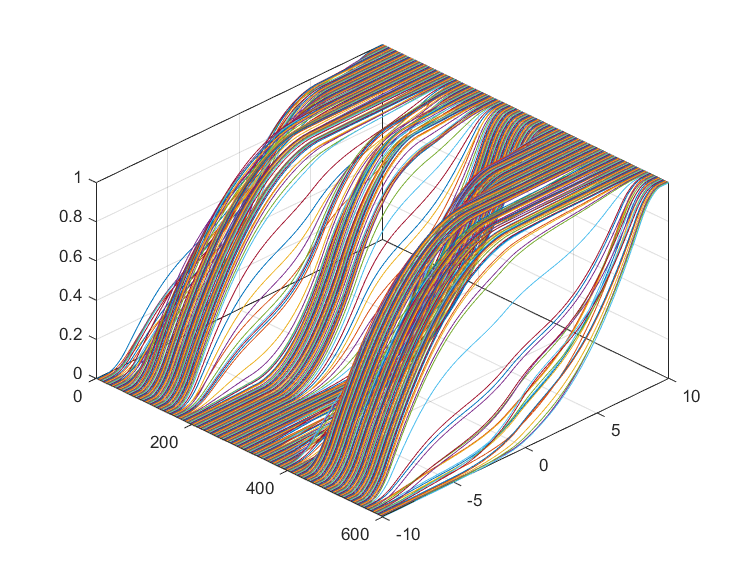}
        \caption{{\small
Empirical distribution functions obtained online as a result of aggregation
of the distributions of three experts by the rule
(\ref{forecast-2b}) for both data generating methods.
}}\label{fig-4}
                  \end{figure}

In this section we present the results of experiments with AA and WA on
synthetic data. The data for experiments were obtained by sampling from a mixture of the
three distinct
probability distributions with the triangular densities.
The time interval is made up of several segments of the same length, and the weights
of the components of the mixture depend on time. We use two methods of mixing of the
three distinct
initial probability distributions.
By Method 1, only one generating probability distribution is a leader
at each segment (i.e., its weight is equal to one). By Method 2, the weights of the mixture
components vary smoothly over time (as shown in section B of Figure~\ref{fig-2}).

Figure~\ref{fig-2} shows the main stages of data mixing
(Method 1 -- left, Method 2 - right) and the results of aggregation of the experts models.  
Section A of the figure shows the realizations of the trajectories of
the three data generating distributions.
The diagram in Section B displays the actual prior probabilities (relative weights)
that were used for mixing of the probability distributions.
Section C shows the result of sampling from the mixture distribution.

There are three experts $i=1,2,3$, which assume that the time series under
study is obtained as a result of sampling from the probability distribution
with the fixed triangular density with given peak and base.
Each expert evaluates the similarity of the testing point of the series
with its distribution using $\CRPS$ score.

We also compare two rules of aggregation of the experts' forecasts, AA (\ref{forecast-2b})
and the weighted average WA (\ref{forecast-2a}).
The diagrams of Sections D and E of Figure~\ref{fig-2} show the weights of the experts
assigned by the corresponding algorithm in the online aggregating process
using rules (\ref{forecast-2b}) and (\ref{forecast-2a}).

Figure~\ref{fig-3} shows the accumulated losses of the experts and
the accumulated losses of the aggregating algorithm for both data generating methods
(Method 1 -- left, Method 2 - right) and for both methods of computing the aggregated
forecasts -- by the rule (\ref{forecast-2b}) and by the rule (\ref{forecast-2a}).
We note an advantage of rule (\ref{forecast-2b})
over the rule (\ref{forecast-2a}) in the case of data generating Method 1,
in which there is a rapid change in leadership of the data generating models.

Figure~\ref{fig-4} shows in 3D format the empirical distribution functions obtained
online by Algorithm 3 for both data generating models and the rule (\ref{forecast-2b}).

\subsection{Probabilistic forecasting of electrical loads}\label{exper-2}

The second group of numerical experiments on probabilistic forecasting
were performed with the data of the 2014 (GEFCOM 2014,Track Load,~\citet{TH2016}).
The time series were divided into training (about 5 years) and testing (about 1 year)
samples.

The main unit of the training sample includes data on hourly electrical load
and data on hourly
temperature measurements for all days of training period. 

\begin{figure}[!htb]

\includegraphics[scale=0.30]{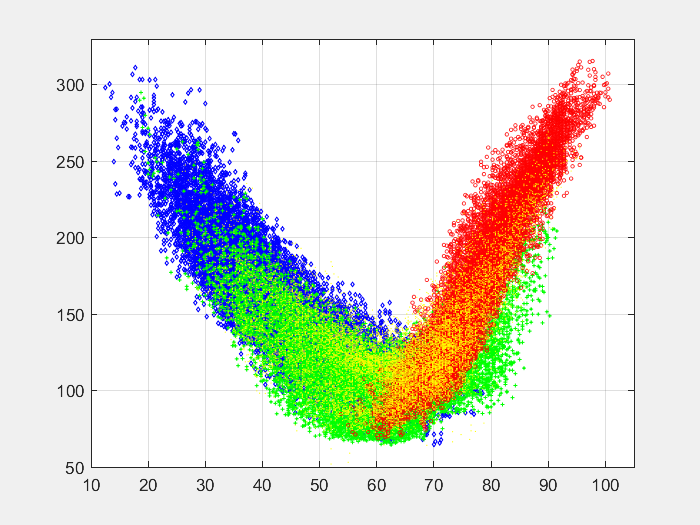}
\includegraphics[scale=0.30]{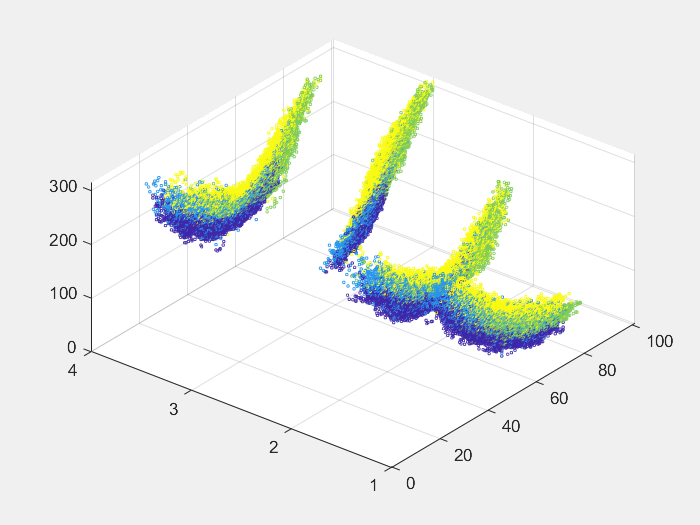}
          \caption{{\small Scatter plots of hourly temperature and electrical loads
          for all days of training period: 
Left figure -- all data marked by seasons;
Right figure -- data grouped by seasons (Winter, Spring, Summer,
Autumn) and time of the day marked in color
(Night, Morning, Day, Evening).
}}\label{fig-5e}
\end{figure}

The training sample shows the dependence of electrical loads on temperature which looks
differently during different seasons and time of the day. Therefore, each expert is trained on
its specific domain where the specific relationship between temperature and electrical load
is observed.
We use the corresponding point clouds of ``temperature--loads'' to define the
probability distribution function of the expert.

The scatter diagrams ``Load - Temperature'' for several sets of calendar parameters
(four seasons of the year and four consecutive intervals of the day, each for 6 hours)
are presented in Figure~\ref{fig-5e}. The diagrams are constructed according to
the training part of the sample.

Figure~\ref{fig-5e} shows the nature of the relationship between potential predictors
and response.
These data show the dependence of electrical loads on temperature.
For each of the scattering diagrams presented, two or three temperature intervals can be
distinguished in such a way that within each interval the point cloud has
a simple ellipsoidal shape. This provides the basis for using a mixture of
normal distributions for the probabilistic forecast of the expected electrical
load according to the short-term temperature forecast.

Scatter patterns on Figure~\ref{fig-5e} can serve as the basis for determining
the pool of the experts. Each of them learns (a predictive probabilistic model)
at sample points related to a predefined calendar segment, for example,
``Winter$\&$Morning'',  etc. These segments should cover all possible combinations of
calendar indicators present in the data.

A set of 21 specialized experts is defined by dividing the calendar space into domains
where the relationship between temperature and electrical load can be described
by a simple and relatively uniform dependence.
To define an expert, a combined sample of historical data consisting of the initial sample
of ``temperature--load'' ensemble, as well as its competence area
(season, time of the day) was determined.
Each expert represents the temperature dependence
of the probabilistic distribution of the magnitude of the electrical load
within a certain domain. These domains represent four daily periods
(morning, afternoon, evening, night)
for each season (winter, spring, summer, autumn). There are 16 such experts in total,
they have numbers 6-21.

The anytime Expert 1 corresponds to the left part of Figure~\ref{fig-5e}, Experts 2-5
correspond to four seasons (see right part of Figure~\ref{fig-5e}). Experts 6-21
correspond to the colored parts of the plots on the right part of Figure~\ref{fig-5e}.
To construct the probability distribution of any expert, we use the method of
Gaussian Mixture Models (GMM), which is applied to the corresponding ensemble
of ``temperature--load''. This probabilistic model of any expert
is presented as a mixture of normal distributions.
The number of components in a Gaussian mixture is preselected (from 1 to 3) depending
on the complexity of the scattering cloud shape for ``temperature--load'' pair
constructed from the training sample.

The main parameter of any expert's model (algorithm) is the temperature forecast. Therefore,
the predictive performance of our algorithm extends as far as the temperature
forecast allows.

\begin{figure}[!htb]
\includegraphics[scale=0.30]{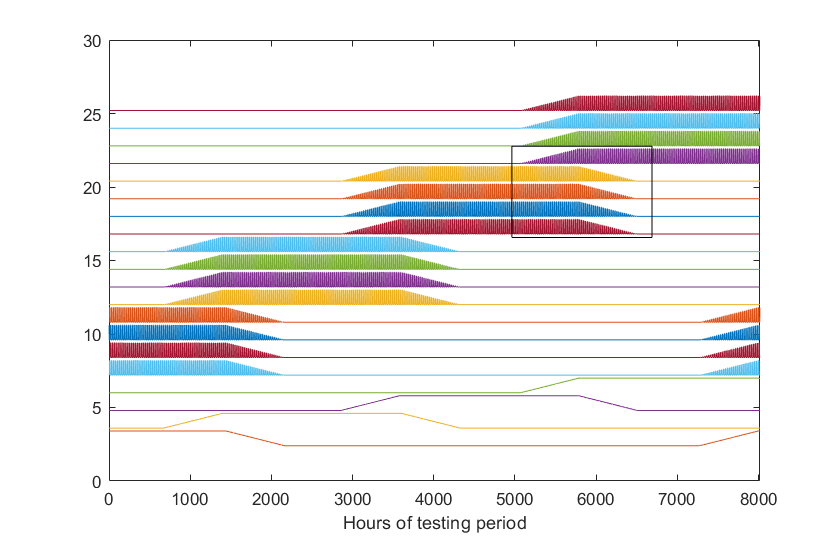}
\includegraphics[scale=0.30]{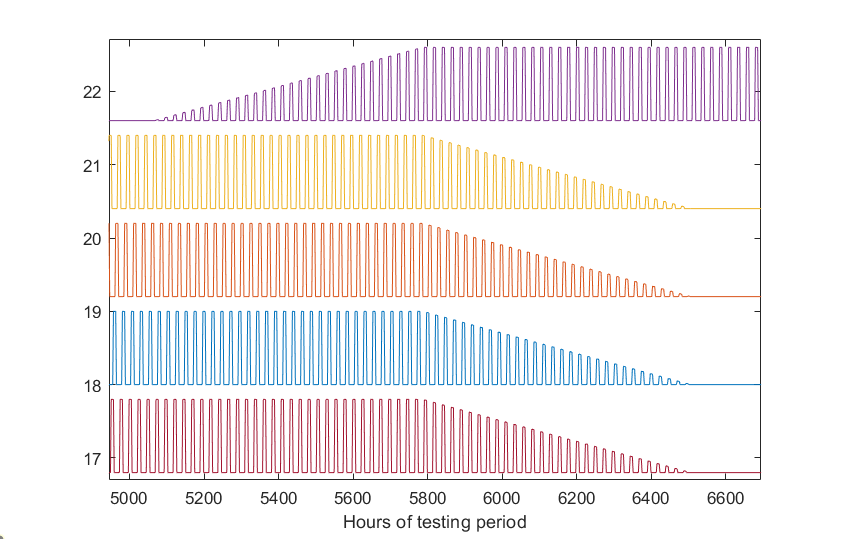}

\caption{{\small
Left part: confidence levels for for Experts 2-5 (season experts) and 6-21
(``season$\&$time of the day''). Right part: enlarged fragment.
Each block is the result of overlaying the confidence levels of the corresponding
seasonal expert with the confidence levels of the day experts. The horizontal axis
displays time, the blocks are vertically spaced.
}}\label{fig-10e}
\end{figure}

\begin{figure}[!htb]
\centering\includegraphics[height=60mm,width=140mm,clip]{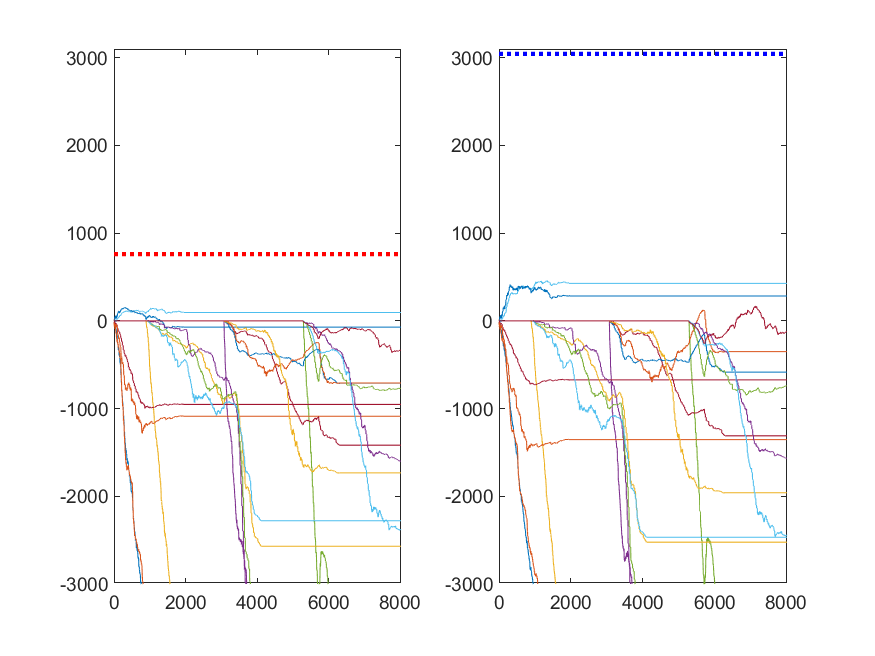}  
          \caption{{\small Discounted regret curves for AA (left) and WA
(right) with respect to each of 21 specialized experts. The dotted lines
above represent the theoretical bounds for the regret.
}}\label{fig-6e}
\end{figure}

\begin{figure}[!htb]
\centering\includegraphics[height=63mm,width=140mm,clip] {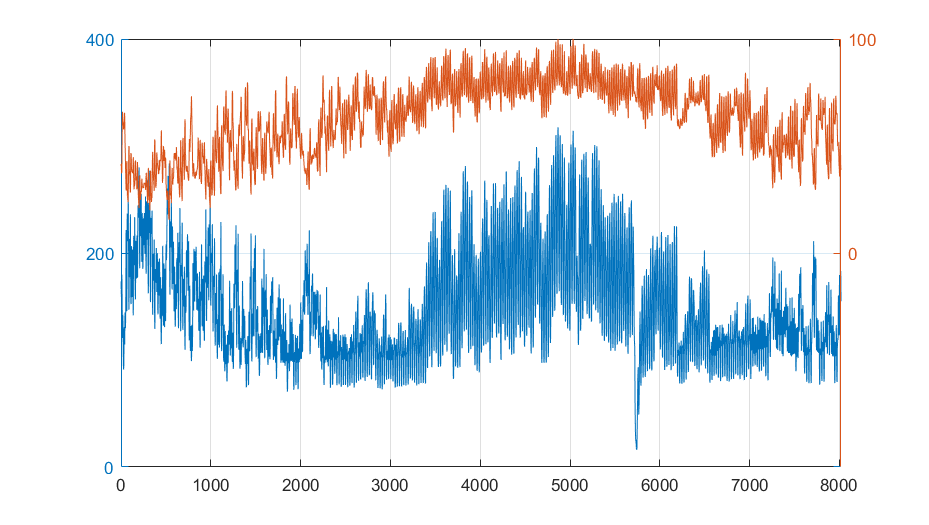}       
          \caption{{\small Temperature (top graph) and hourly electrical loads
(bottom graph) for the testing period. The left vertical axis is the load value, the right vertical
axis is the temperature in Fahrenheit scale.
There is a jump of consumption between 5000 and 6000 hours of testing period, which is then reflected in the results
of the forecasting algorithms.
}}\label{fig-1}
\end{figure}

\begin{figure}[!htb]
\includegraphics[height=47mm,width=67mm,clip]{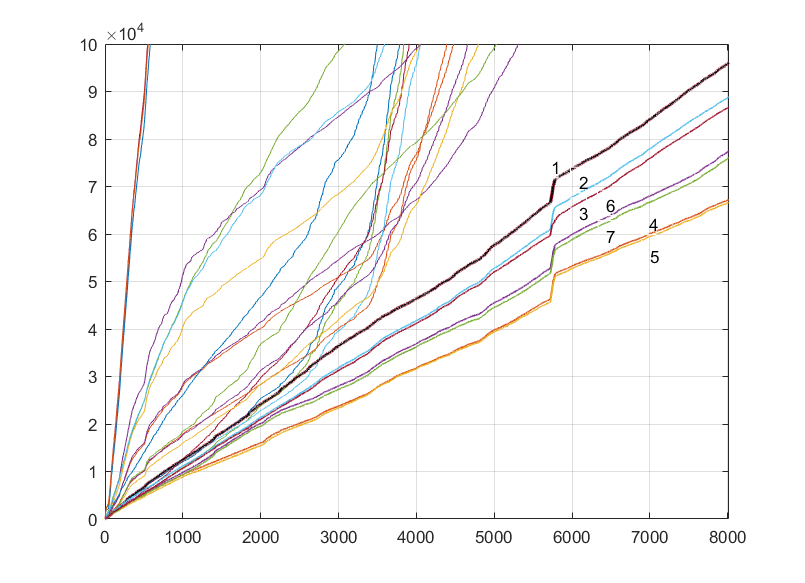}
\includegraphics[height=47mm,width=65mm,clip]{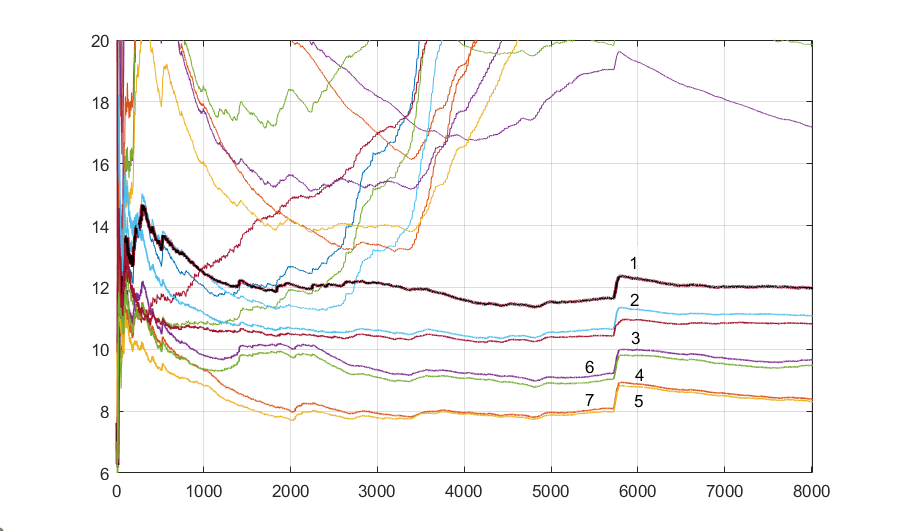}
          \caption{{\small {\bf Comparative study
of learning with/without specialization of the experts.}
Accumulated losses (left) and their time averages (right):
of all 21 specialized experts working any time
(there is some difference with curves on Figure~\ref{fig-6e},
where the discounted regrets are presented);
1-- losses of the anytime expert trained on the entire sample;
2 and 3 -- results of aggregation by WA and AA,
where confidence levels of the experts are set to 1;
4-5 -- results of aggregation by WA and AA algorithms
using non-trivial overlapping smooth confidence levels;
6-7 -- the same for the case where the expertise areas of the experts
do not overlap (sleeping and non-sleeping experts).
{AA is always slightly outperforms WA.
}}}\label{fig-7e}
\end{figure}

In the experiments, which are presented in Figures~\ref{fig-10e}--\ref{fig-7e},
a particular forecasting problem is considered, that is the short-term forecasting of
a probability distribution function for one hour in advance. We use the current
temperature as its forecast on one hour ahead.

Each expert is trained on its specific domain of time interval.
The scope of each expert is determined by its confidence values.
Moving from one domain to another, an expert, which was tuned to the previous
domain, gradually loses his predictive abilities. To take this into account,
when forecasting, we define a smooth extension of the domain of any expert.
Inside the area for which the expert was tuned,
its confidence values are equal
to 1, and outside this area they decrease linearly from 1 to 0; moreover, the area
of decrease for a seasonal expert is equal to half of the duration of the season,
and the area of decrease for a daily expert is equal to two hours (the specific domain
of any daily expert is equal to six hours).

Confidence levels of Seasonal Experts 2-5, as well as corresponding Experts 6-21,
are presented as blocks on Figure~\ref{fig-10e}.
Each block is the result of overlaying the confidence levels of the corresponding
seasonal expert with the confidence levels of the
experts.\footnote{These values are simply multiplied.}

The constructed experts and methods of their aggregation were tested on the testing sample.
Temperature and hourly electrical loads for the testing period are presented
on Figure~\ref{fig-1}.

When forecasting, the smooth areas of expert
competence are chosen wider than those areas in which these experts were trained.
Thus, each expert competes with other experts working at overlapping intervals
using the corresponding algorithm for combining experts with confidence
levels from Section~\ref{CRPS-3},
like it was done for computing the pointwise forecasts by~\citet{VyT2019-1}.

\begin{figure}[!htb]
\includegraphics[height=60mm,width=140mm,clip]{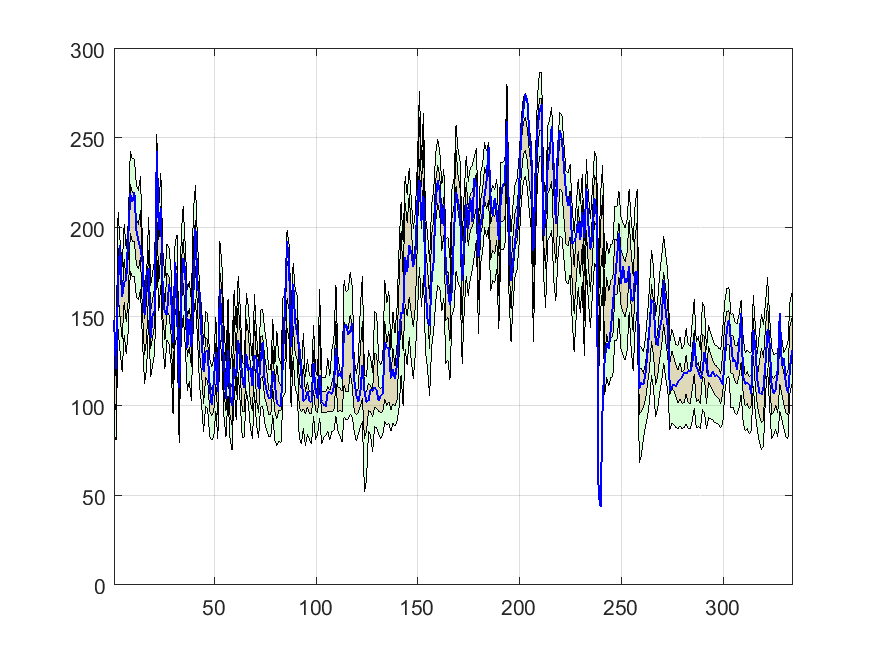}
          \caption{{\small Interquantile intervals (semitones) and 
actual loads (blue line) at 12 o'clock (noon).}}\label{fig-8e}
\end{figure}

\begin{figure}[!htb]
\includegraphics[height=60mm,width=140mm,clip]{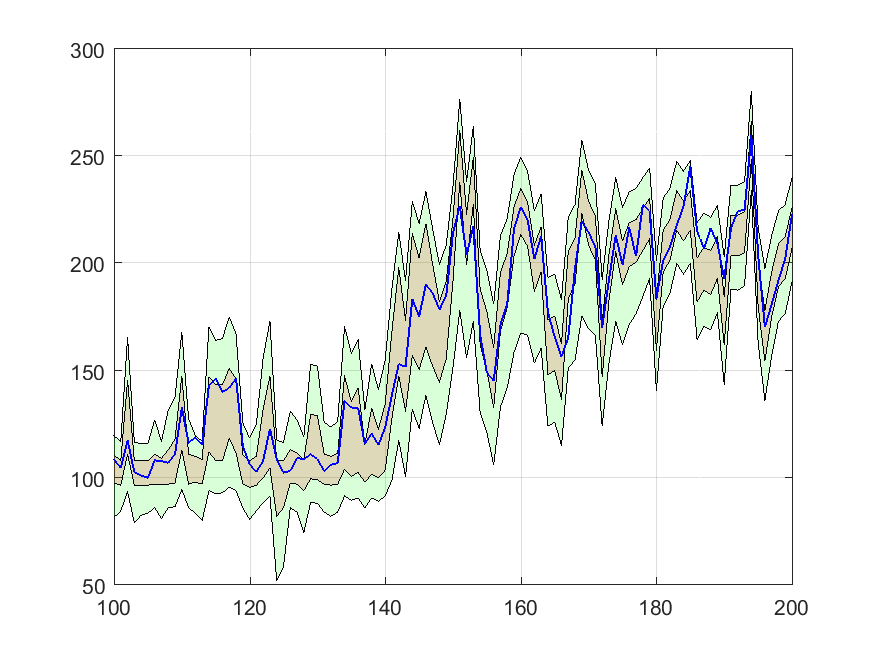}
          \caption{{\small Quantile spacing (semitones)
$[0.25,0.75]$ and $[0.05,0.95]$ and current
loads at 12 o'clock - noon (solid line). Fragment of Fig.~\ref{fig-8e}
from March 10, 2011 to July 20 of the same year. 
}}\label{fig-8ee}
\end{figure}

The regret curves $T\to\sum\limits_{t=1}^T p_{i,t}(h_t-l_{i,t})$
for AA and WA with respect to each of 21 specialized experts are presented in
Figure~\ref{fig-6e}.
The dotted lines above represent the theoretical bounds
for the regret (see the inequality~(\ref{mTT-1afaa})).

Two ways of aggregation of the experts by AA and WA were tested.
In the first method of aggregation, confidence levels of all experts were equal to 1.
In the second way, algorithms AA and WA use specialized experts,
where theirs confidence levels are set externally.
Non-zero confidences correspond to the training intervals of specialized experts,
but are somewhat wider and monotonically decrease to zero outside these intervals
(see example in Figure~\ref{fig-10e}).

To justify the role of confidence parameters, the comparative experiments were conducted.
Their results are presented in Figure~\ref{fig-7e}.
The accumulated losses and their time averages
are presented in Figure~\ref{fig-7e}.
These curves show that specialized experts, which were trained
only for certain types of data, quickly lose their effectiveness in other types
of data areas and generally suffer large losses. An exception is Expert 1, which
was trained on all types of data, but the aggregating algorithms AA and WA with confidence
essentially outperform it.

Other experiments study the effects of smooth and constant confidence levels.
During the first experiment, all confidence values for each expert were equal to 1:
curves 2 and 3 (in Figure~\ref{fig-7e}) represent results of their aggregation by AA and WA,
where confidence levels of the experts are set to 1.
In the second experiment, AA and WA algorithms used the experts predictions within
the levels of their confidence: curves 4-5 represent results of aggregation by WA and AA
algorithms using non-trivial overlapping smooth confidence levels.

We also test the binary case, where confidence levels of the experts take only values 0 or 1
(sleeping and non-sleeping experts): curves 6-7 represent results of aggregation
by WA and AA for the binary case where the expertise areas of the experts do not overlap.

The results of the experiments show that the use of smooth confidence levels
of specialized experts increases the efficiency of the process of online adaptation
compared to those cases where confidence values are binary or when they are not used at all
(when they are always equal to 1).

These results also show that AA in all experiments outperforms WA.

Examples of hourly forecasts are shown in Fig.~\ref{fig-8e} and~\ref{fig-8ee}.

\section{Conclusion}

In this paper, the problem of aggregating the probabilistic forecasts is considered.
In this case, Continuous Ranked Probability Score ($\CRPS$) is a popular among practitioners
example of proper scoring rule for continuous outcomes.
We incorporate this loss function in PEA framework and
present its theoretical analysis. We have proved that the $\CRPS$ loss function is
mixable. This implies that all machinery of the Vovk aggregating algorithm can be
applied to this loss function.
Basing on mixability of $\CRPS$, we analyze two methods for calculating the predictions
using the aggregating algorithm (AA) and the weighted average of
forecasts of the experts (WA).
The time-independent upper bounds for the regret were obtained for both methods.

We illustrate the theoretical results with computer experiments.
In Section~\ref{exper-1} we test the performance of two methods of aggregation,
AA and WA, on synthetic data. We use three probabilistic models for generating data.
The same models are used as experts. Our experiments show how quickly
the mixing algorithms can adapt to the data generation strategy (see Figure~\ref{fig-2}).

These results show that two methods of computing
forecasts AA and WA lead to similar empirical cumulative losses while the
rule (\ref{forecast-2b}) for AA results in four times less regret bound than
(\ref{forecast-2a}) for WA. We note a significantly better performance of method AA
over method WA (\ref{forecast-2a}) in the case where
there is a rapid change in leadership of the data generating models.

We have incorporated a smooth generalization of the method of specialized experts into
the aggregating algorithm, which allows us to combine the probabilistic
predictions of the specialized experts with overlapping domains of theirs competence.

This paper applies our approach to a popular problem of
predicting electricity consumption using Gaussian mixture models as experts.
We propose a technology for developing specialized experts and
learning their probability distributions using ensembles of learning samples.

A set of 21 specialized experts is defined by dividing the calendar space into domains
where the relationship between temperature and electrical load can be described
by a simple and relatively uniform dependence.
The main parameter of any expert's model (algorithm) is the temperature forecast.
Therefore, the predictive performance of our algorithm extends as far as the temperature
forecast allows.
The problem of predicting temperature for several hours in advance is beyond the scope
of this study and is a separate problem that may be the subject of future research.

The results of these experiments show that the use of smooth confidence levels
of specialized experts increases the efficiency of the process of online adaptation
compared to those cases where confidence values are binary or when they are not
used at all.

The proposed methods are closely related to the so called ensemble forecasting
(\citet{TMB2017}).
In practice, the output of physical process models are usually not probabilities,
but rather ensembles. Ensemble forecasts are based on a set of physical models.
Each model may have its own physical formulation, numerical formulation and input data.
An ensemble is a collection of model trajectories
generated using different initial conditions of model equations.
Consequently, the individual ensemble members represent likely scenarios of the future
physical system development, consistent with the currently available incomplete
information. It is possible to apply the aggregation methods developed directly to
the data represented in the form of ensembles.

\section*{Acknowledgements}

This paper is an extended version of COPA 2019
(Conformal and Probabilistic Prediction with Applications) paper by~\citet{VyT2019-2}.
This work was partially supported by the
Russian Foundation for Basic Research, project 20-01-00203.

The authors are grateful to Vladimir Vovk and Yuri Kalnishkan for useful discussions.
The authors thank the anonymous reviewers, whose comments significantly
improved the presentation of this work.

\appendix

\section{Auxiliary results}

\subsection{Regret analysis for AA}\label{regret-ana-1}

Assume that a loss function $\lambda(f,y)$ is $\eta$-mixable. Let
$\w^*_t=(w^*_{1,t},\dots ,w^*_{N,t})$ be the normalized weights and
$\f_t=(f_{1,t},\dots ,f_{N,t})$ be the experts' forecasts
at step $t$. Define in Protocol 1 the learner's forecast $f_t=\Subst(\f_t,\w^*_t)$.
By (\ref{mix-1}) $\lambda(f_t,y_t)\le g_t(y_t)$ for all $t$, where
$g_t(y)$ is defined by (\ref{superpred-1}).

Let $H_T=\sum\limits_{t=1}^T \lambda(f_t,y_t)$ be the accumulated loss of the
learner and $L^i_T=\sum\limits_{t=1}^T \lambda(f_{i,t},y_t)$ be the
accumulated loss of an expert $i$. By definition
$g_t(y_t)=-\frac{1}{\eta}\ln\frac{W_{t+1}}{W_t}$, where
$W_t=\sum\limits_{i=1}^N w_{i,t}$ and $W_1=1$.
By the weight update rule (\ref{wei-up-1}), we obtain
$w_{i,t+1}=\frac{1}{N}e^{-\eta L^i_t}$.

By telescoping, we obtain the time-independent bound
\begin{eqnarray}
H_T\le\sum\limits_{t=1}^T g_t(y_t)=-\frac{1}{\eta}\ln W_{T+1}\le
L^i_T+\frac{\ln N}{\eta}
\label{prop-1}
\end{eqnarray}
for any expert $i$ regardless of which sequence
of outcomes is observed.

\subsection{Proof of Lemma~\ref{Adam-1}}\label{adam-2}

{\it Proof.} Let the forecasts $\c_i=(c^1_i,\dots, c^d_i)$ of the experts $1\le i\le N$
and a probability distribution $\p=(p_1,\dots ,p_N)$ on the set of the experts be given.

Since the loss function $\lambda(f,y)$ is $\eta$-mixable, we can apply the
aggregation rule to each $s$th column $\e^s=(c^s_1,\dots ,c^s_N)$ of coordinates separately:
define $f^s=\Subst(\e^s,\p)$ for $1\le s\le d$.
Rewrite the inequality (\ref{mix-1}):
\begin{eqnarray}
e^{-\eta\lambda(f^s,y)}\ge\sum\limits_{i=1}^N
e^{-\eta\lambda(c^s_i,y)}p_i
\label{superpred-1n}
\end{eqnarray}
for $1\le s\le d$ and for any $y$.

Let $\y=(y^1,\dots ,y^d)$ be a vector of outcomes.
Multiplying the inequalities (\ref{superpred-1n}) for $s=1,\dots ,d$
and $y=y^s$, we obtain
\begin{eqnarray}
e^{-\eta\sum_{s=1}^d\lambda(f^s,y^s)}
\ge\prod_{s=1}^d\sum_{i=1}^N e^{-\eta\lambda(c^s_i,y^s)}p_i.
\label{aggr-rule-1fuapp}
\end{eqnarray}

The generalized H\"older inequality says that
$$
\|G_1G_2\cdots G_d\|_r\le\|G_1\|_{q_1}\|G_2\|_{q_2}\cdots\|G_d\|_{q_d},
$$
where $\frac{1}{q_1}+\dots +\frac{1}{q_d}=\frac{1}{r}$,  $q_s\in (0,+\infty)$
and $G_s\in L^{q_s}$ for $1\le s\le d$ (\citet{Loe77}).
Let $q_s=1$ for all $1\le s\le d$, then $r=1/d$.
Let $G_s(i)=e^{-\eta\lambda(c^s_i,y^s)}$ for $s=1,\dots ,d$ and
$\|G_s\|_1=E_{i\sim \p}[G_s(i)]=\sum\limits_{i=1}^N G_s(i)p_i$.
Then using the inequality (\ref{aggr-rule-1fuapp}), we obtain
\begin{eqnarray*}
e^{-\eta\sum_{s=1}^d\lambda(f^s,y^s)}
\ge\left(\sum_{i=1}^N e^{-\eta\frac{1}{d}\sum\limits_{s=1}^d\lambda(c^s_i,y^s)}p_i\right)^d.
\end{eqnarray*}
or, equivalently,
\begin{eqnarray}
e^{-\frac{\eta}{d}\lambda(\f,\y)}\ge\sum_{i=1}^N
e^{-\frac{\eta}{d}\lambda(\c_i,\y)}p_i
\label{ii-1}
\end{eqnarray}
for all $\y=(y^1,\dots ,y^d)$, where $\f=(f^1,\dots ,f^d)$.

The inequality (\ref{ii-1}) means that the generalized loss function
$\lambda(\f,\y)$ is $\frac{\eta}{d}$-mixable.

By (\ref{wei-up-1}), the weights update rule for generalized loss function
in Protocol 1 is
\begin{eqnarray*}
w_{i,t+1}=w_{i,t}e^{-\frac{\eta}{d}\lambda(\f_{i,t},\y_t)}\mbox{ for } t=1,2,\dots,
\label{wei-up-1m}
\end{eqnarray*}
where $\eta>0$ is a learning rate for the source function. The normalized weights
$\w^*_t=(w^*_{i,t},\dots, w^*_{i,t})$ are defined by (\ref{weight-update-1}).
At any round $t$, the learner forecast $\f_t=(f^1_t,\dots ,f^d_t)$
is defined as $f^s_t=\Subst(\e^s_t,\w^*_t)$ for each $s=1,\dots ,d$,
where $\e^s_t=(f^s_{1,t},\dots ,f^s_{N,t})$.
$\Box$

\subsection{Mixing past posteriors}\label{mix-past-1}

\begin{table}
\caption{Some values of the parameter $\alpha$ and the corresponding accumulated losses of Algorithm 3,
when the first synthetic data generation model (Method 1) was used.
The values of the losses are normalized relative to the losses of the algorithm WA 
for $\alpha=0$.}

\label{table-1} \vskip 0.15in
\begin{center}
\begin{small}
\begin{sc}
\begin{tabular}{|l|l|l|l|l|l|l|l|l|l|l|}
\hline
$\alpha$  & 0 & 0.0001 & 0.001 & 0.005 &  0.01 & 0.05 & 0.1 & 0.2\\
\hline
AA  & 0.984 & 0.596 &  0.542 & 0.513 & 0.508 & 0.564 & 0.657 & 0.824\\
\hline
WA  &  1.000 & 0.958 &  0.869 &  0.759 & 0.728 &  0.816 & 0.957 &  1.115\\
\hline

\end{tabular}%
\end{sc}
\end{small}
\end{center}
\vskip -0.1in
\end{table}

We have used mixing past posteriors modification of Algorithms 3 and 3a
(see Figure 1 and mixing scheme Fixed Share Update
(to start vector) on Table 1 by~\citealt{BoW2002}),
where the rules (\ref{weight-up-3}) and (\ref{for-1b-2c}) are replaced with
\begin{eqnarray*}
w_{i,t+1}=\frac{\alpha}{N}+(1-\alpha)\frac{\tilde w_{i,t}}{\sum\limits_{j=1}^N\tilde w_{j,t}},\mbox{ where }~~~~~~~~~
\\
\tilde w_{i,t}=w_{i,t}e^{-\eta (p_{i,t}\CRPS(F_{i,t},y_t)+(1-p_{i,t})\CRPS(F_t,y_t))}.
\end{eqnarray*}
The value of parameter $\alpha$ was not optimized.
Some values of the parameter $\alpha$ and the corresponding accumulative losses of Algorithm 3
for the first synthetic data generation model (Method 1) are presented on Table~A.1

The loss values given in the table show that in this particular case, a significant decrease
in losses occurs already at the first nonzero value of the parameter $\alpha$.
There is a jump in losses at the first nonzero tested value $\alpha = 0.0001$,
after which their change was insignificant. We have chosen the value $\alpha=0.001$ within
the interval of relative stabilization of the corresponding losses.
Optimization of the parameter value $\alpha$ can serve as a subject for further research.

\bibliographystyle{elsarticle-num-names}
\bibliographystyle{elsarticle}

\end{document}